\documentclass[11pt,letterpaper]{article}
\usepackage{ijcnlp2017}
\usepackage{times}
\usepackage{latexsym}
\usepackage{graphicx}
\usepackage{subfigure}

\ijcnlpfinalcopy

\graphicspath{{figures/}}

\title{Reusing Neural Speech Representations for Auditory Emotion Recognition}

\author{Egor Lakomkin \hspace{0.5cm} Cornelius Weber\hspace{0.5cm} Sven Magg\hspace{0.5cm} Stefan Wermter \\
  Department of Informatics, Knowledge Technology\\ University of Hamburg \\ Vogt-Koelln Str. 30, 22527 Hamburg, Germany \\
  {\tt \{lakomkin, weber, magg, wermter\}@informatik.uni-hamburg.de}
  }
\date{}

\begin{document}

\maketitle

\begin{abstract}
  Acoustic emotion recognition aims to categorize the affective state of the speaker and is still a difficult task for machine learning models. The difficulties come from the scarcity of training data, general subjectivity in emotion perception resulting in low annotator agreement, and the uncertainty about which features are the most relevant and robust ones for classification. In this paper, we will tackle the latter problem. Inspired by the recent success of transfer learning methods we propose a set of architectures which utilize neural representations inferred by training on large speech databases for the acoustic emotion recognition task. Our experiments on the IEMOCAP dataset show ~10\% relative improvements in the accuracy and F1-score over the baseline recurrent neural network which is trained end-to-end for emotion recognition.
\end{abstract}

\section{Introduction}

Speech emotion recognition (SER) has received growing interest and attention in recent years. Being able to predict the affective state of a person gives valuable information which could improve dialog systems in human-computer interaction. To fully understand a current emotion expressed by a person, also knowledge of the context is required, like facial expressions, the semantics of a spoken text, gestures and body language, and cultural peculiarities. This makes it challenging even for people that have all this information to accurately predict the affective state. In this work, we are focusing solely on inferring the speaker's emotional state by analysing acoustic signals which are also the only source of information in situations when the speaker is not directly observable. \par

\begin{figure}[t]
  \includegraphics[width=\linewidth]{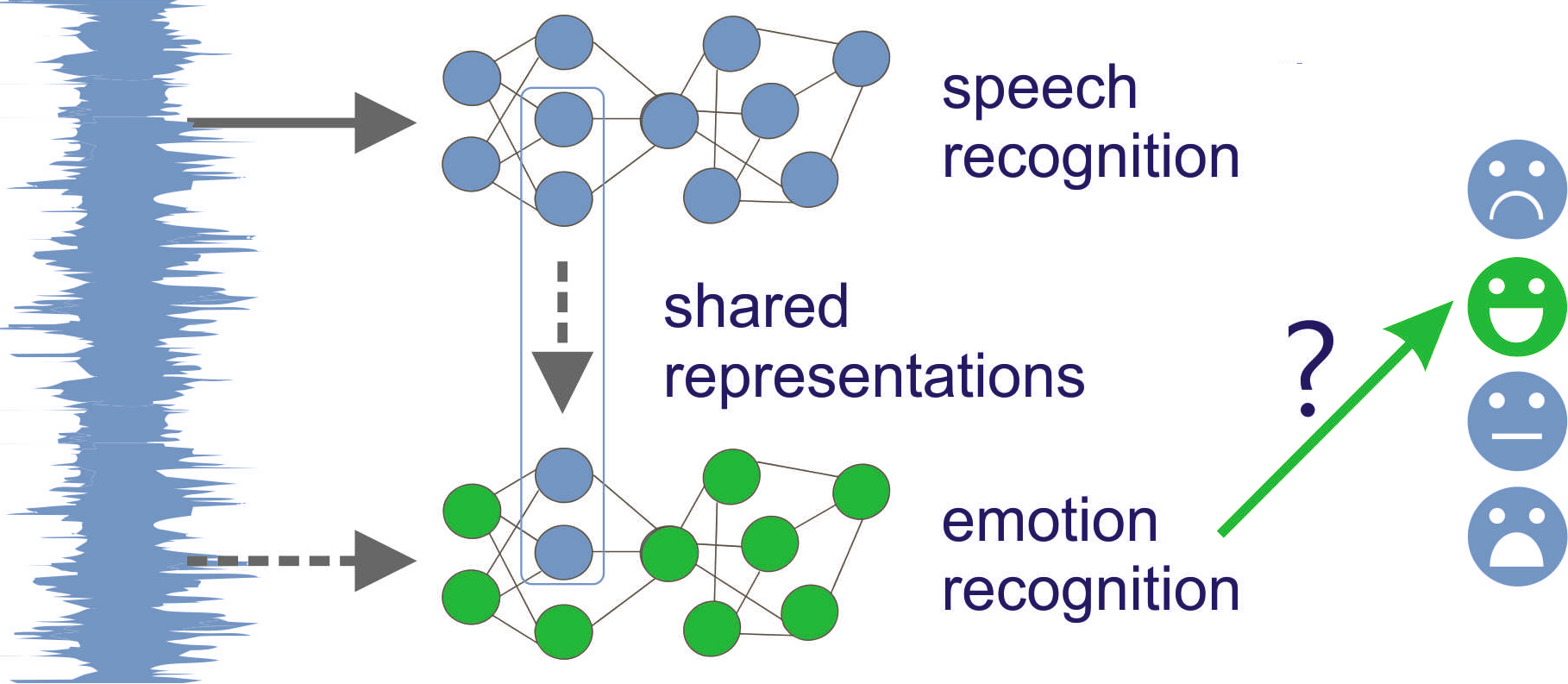}
  \caption{High-level system architecture. Acoustic emotion recognition system uses neural speech representations for affective state classification. }
  \label{fig:system_main}
\end{figure}

    With recent advances in deep learning, which made it possible to train large end-to-end models for image classification \cite{simonyan_very_2014}, speech recognition \cite{hannun_deep_2014} and natural language understanding \cite{sutskever_sequence_2014}, the majority of the current work in the area of acoustic emotion recognition is neural network-based. Diverse neural architectures were investigated  based on convolutional and recurrent neural networks \cite{fayek_evaluating_2017,trigeorgis_adieu_2016}. Alternatively, methods based on linear models, like SVM with careful feature engineering, still show competitive performance on the benchmark datasets \cite{schuller2013interspeech, schuller2009acoustic}. Such methods were popular in computer vision until the AlexNet approach \cite{krizhevsky_imagenet_2012} made automatic feature learning more wide-spread. The low availability of annotated auditory emotion data is probably one of the main reasons for traditional methods being competitive. The appealing property of neural networks compared to SVM-like methods is their ability to identify automatically useful patterns in the data and to scale linearly with the number of training samples. These properties drive the research community to investigate different neural architectures. In this paper, we present a model that neither solely learns feature representations from scratch nor uses complex feature engineering but uses the features learned by a speech recognition network. \par
    Even though the automatic speech recognition (ASR) task is agnostic to the speaker's emotion and focuses only on the accuracy of the language transcription, low-level neural network layers trained for ASR might still extract useful information for the task of SER. Given that the size of the data suitable for training ASR systems is significantly larger than SER data, we can expect that trained ASR systems are more robust to speaker and condition variations.  Recently, a method of transferring knowledge learned by the neural network from one task to another has been proposed  \cite{rusu_progressive_2016,anderson_beyond_2016}. This strategy also potentially prevents a neural network from overfitting to the smaller of the two datasets and could work as an additional regularizer. Moreover, in many applications, such as dialog systems, we would need to transcribe spoken text and identify its emotion jointly. \par
    
In this paper, we evaluate several dual architectures which integrate representations of the ASR network: a fine-tuning and a progressive network. The fine-tuning architecture reuses features learnt by the recurrent layers of a speech recognition network and can use them directly for emotion classification by feeding them to a softmax classifier or can add additional hidden SER layers to tune ASR representations. Additionally, the ASR layers can be static for the whole training process or can be updated as well by allowing to backpropagate through them. The progressive architecture complements information from the ASR network with SER representations trained end-to-end. Therefore, in contrast to the fine-tuning model, a progressive network allows learning such low-level emotion-relevant features which the ASR network never learns since they are irrelevant to the speech recognition task. Our contribution in this paper is two-fold: 1) we propose several neural architectures allowing to model speech and emotion recognition jointly, and 2) we present a simple variant of a fine-tuning and a progressive network which improves the performance of the existing end-to-end models.

\begin{figure}
  \includegraphics[scale=0.43]{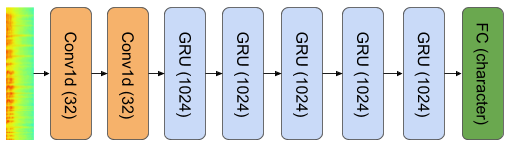}
  \caption{Architecture of the ASR model used in this work, following the DeepSpeech 2 architecture.  }
  \label{fig:asrdeepspeech}
\end{figure}

\section{Related work}

The majority of recent research is aimed at searching for optimal neural architectures that learn emotion-specific features from little or not processed data. An autoencoder network \cite{ghosh_representation_2016} was demonstrated to learn to compress the speech frames before the emotion classification.  An attention mechanism \cite{huang_attention_2016} was proposed to adjust weights for each of the speech frames depending on their importance. As there are many speech frames that are not relevant to an expressed emotion, such as silence, the attention mechanism allows focusing only on the significant part of the acoustic signal. Another approach probabilistically labeling each speech frame as emotional and non-emotional was proposed by \cite{DBLP:journals/corr/ChernykhSP17}. A combination of convolutional and recurrent neural networks was demonstrated by \cite{trigeorgis_adieu_2016} by training a model directly from the raw, unprocessed waveform, significantly outperforming manual feature engineering methods.  \par
    One of the first examples of the knowledge transfer among neural networks was demonstrated by \cite{bengio_deep_2012} and \cite{yosinski_how_2014}. Eventually, fine-tuning became the de-facto standard for computer vision tasks that have a small number of annotated samples, leveraging the ability of trained convolutional filters to be applicable to different tasks. Our work is mainly inspired by recently introduced architectures with an ability to transfer knowledge between recurrent neural networks in a domain different from computer vision \cite{rusu_progressive_2016, anderson_beyond_2016}.\par
    
Previously, to our knowledge, there was only one attempt  to analyze the correlation between automatic speech and emotion recognition \cite{fayek_correlation_2016}. This approach showed the possibility of knowledge transfer from the convolutional neural acoustic model trained on the TIMIT corpus \cite{garofolo_darpa_1993} for the emotion recognition task. The authors proposed several variants of fine-tuning. They reported a significant drop in the performance by using the ASR network as a feature extractor and training only the output softmax layer, compared to an end-to-end convolution neural network model. Gradual improvements were observed by allowing more ASR layers to be updated during back-propagation but, overall, using ASR for feature extraction affected the performance negatively.

\section{Models and experiment setup}

\subsection{Models}
\par We introduce two models that use a pre-trained ASR network for acoustic emotion recognition. The first one is the fine-tuning model which only takes representations of the ASR network and learns how to combine them to predict an emotion category. We compare two variants of tuning ASR representations: simply feeding them into a softmax classifier or adding a new Gated Recurrent Units (GRU) layer trained on top of the ASR features. The second is the progressive network which allows us to train a neural network branch parallel to the ASR network which can capture additional emotion-specific information. We present all SER models used in this work in Figure \ref{fig:systems}.

\subsubsection{ASR model}

Our ASR model (see Figure \ref{fig:asrdeepspeech}) is a combination of convolutional and recurrent layers inspired by the DeepSpeech \cite{hannun_deep_2014} architecture for speech recognition. Our model contains two convolutional layers for feature extraction from power FFT spectrograms, followed by five recurrent bi-directional GRU layers with the softmax layer on top, predicting the character distribution for each speech frame (ASR network in all our experiments, left branch of the network in the Figures \ref{fig:systems}b, \ref{fig:systems}c  and \ref{fig:systems}d). The ASR network is trained on pairs of utterances and the corresponding transcribed texts (see 3.2.2 ``Speech data`` section for details). Connectionist Temporal classification (CTC) loss \cite{graves_connectionist_2006} was used as a metric to measure how good the alignment produced by the network is compared to the ground truth transcription.  Power spectrograms were extracted using a Hamming window of 20ms width and 10 ms stride, resulting in 161 features for each speech frame. We trained the ASR network with Stochastic Gradient Descent with a learning rate of 0.0003 divided by 1.1 after every epoch until the character error rate stopped improving on the validation set (resulting in 35 epochs overall).  In all our experiments we keep the ASR network static by freezing its weights during training for SER.

\subsubsection{Baseline}

As a baseline (see Figure \ref{fig:systems}a), we used a two-layer bi-directional GRU neural network. Utterances were represented by averaging hidden vector representations obtained on the second GRU layer and fed to a softmax layer for emotion classification. Dropout with the probability of 0.25 was applied to the utterance representation during training to prevent overfitting. We evaluate this architecture as a baseline as it was proven to yield strong results on the acoustic emotion recognition task \cite{huang_attention_2016}. As there are significantly less emotion-annotated samples available the SER-specific network is limited to two layers compared to the ASR network.

\subsubsection{Fine-tuning model}
\begin{figure*}[t]
  \includegraphics[width=16cm]{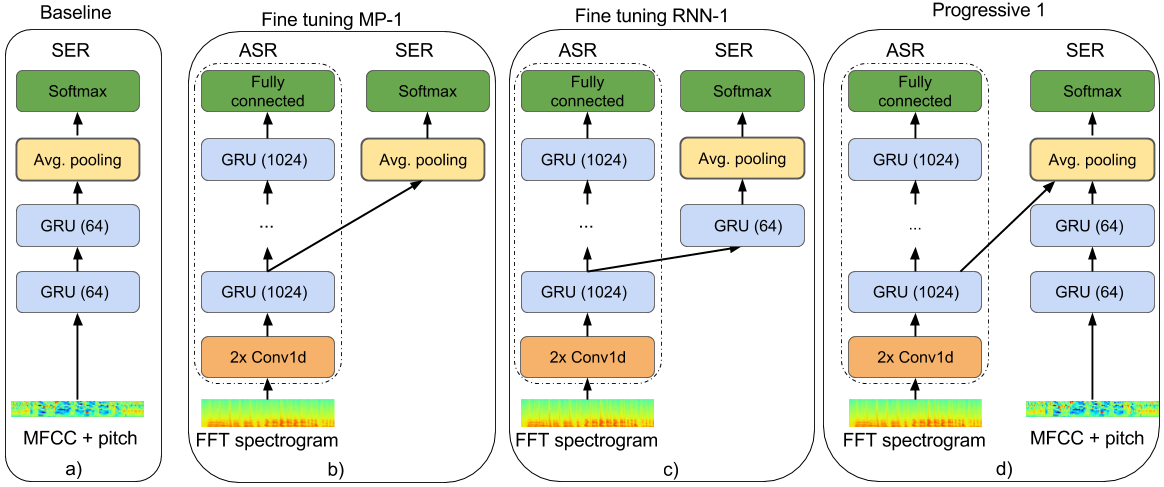}
  \caption{(a) Baseline speech emotion recognition (SER) model, containing two bi-directional GRU layers (b) A variant of the fine-tuning network (”Fine-tuning MP-1”) which uses the temporal pooled representations of the first recurrent layer of the ASR network. (c) A variant of the fine-tuning network (”Fine-tuning RNN-1”) which uses hidden memory representations of the first recurrent layer of the ASR network as input to a new emotion-specific GRU layer. (d) The progressive network combines representations from the emotion recognition path with
the temporal pooled first layer representations of the ASR network (here ”Progressive net 1”). A concatenated vector is fed into the softmax layer for the final emotion classification. The ASR branch of the network remains static and is not tuned during backpropagation in (b), (c) and (d). }
  \label{fig:systems}
\end{figure*}

We propose several variants of reusing speech representations: 1) By averaging hidden memory representations of the layer number $x$ of the ASR network (\textit{Fine-tuning MP-$x$} later in the  text where MP stands for \textit{Mean Pooling}), we train only the output softmax layer to predict an emotion class. (2) We feed hidden memory representations as input to a new GRU network (emotion-specific) initialized randomly and trained for emotion classification (\textit{Fine-tuning RNN-$x$}). The intuition is that bottom layers of the ASR network can be used as feature extractors and the top level GRU can combine them to predict the emotion class. Similar to the \textit{Fine-tuning MP-$x$} setup we average representations of the newly attached GRU layer and feed them to the classifier. 
  In both experiments, dropout with the rate of 0.25 was applied to averaged representations.   Figure \ref{fig:systems}b shows the \textit{Fine-tuning MP-1} model pooling ASR representation of the first ASR layer, and Figure \ref{fig:systems}c shows the \textit{Fine-tuning RNN-1} setup.

\subsubsection{Progressive neural network}

The progressive network combines fine-tuning with end-to-end training. The scheme of the model is presented in Figure \ref{fig:systems}d. It contains two branches: first, a speech recognition branch (left in Figure \ref{fig:systems}d and identical to Figure \ref{fig:asrdeepspeech}) which is static and not updated, and second, an emotion recognition branch (right), with the same architecture as the baseline model which we initialized randomly and trained from scratch.  We feed the same features to the emotion recognition branch as to the baseline model for a fair comparison. Theoretically, a network of such type can learn task specific features while incorporating knowledge already utilized in the ASR network if it contributes positively to the prediction.

\begin{figure*}[t]
  
  \centering
  \includegraphics[height=8cm, width=16cm]{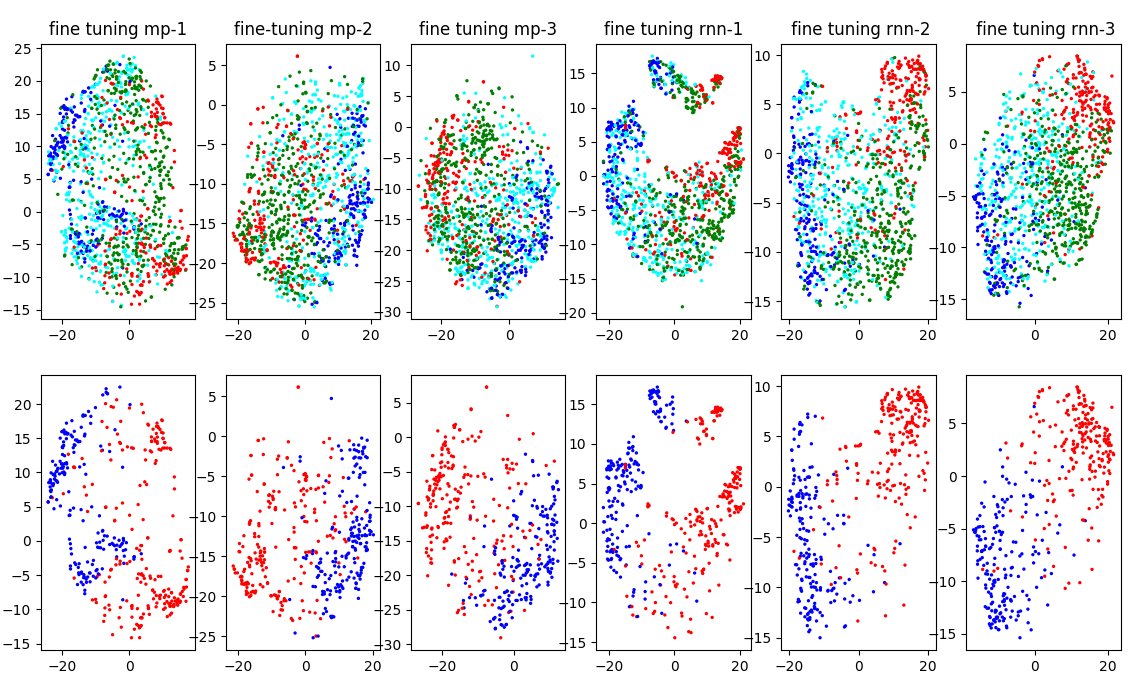}
  \caption{Representations of the IEMOCAP utterances generated by the \textit{Fine-tuning MP-$x$} and \textit{Fine-tuning RNN-$x$} networks ($x$ stands for the ASR layer number of which the representation is used) projected into 2-dimensional space using the t-SNE technique. Top: all four classes, bottom: only \textit{Sadness} and \textit{Anger}. Color mapping: \textit{Anger} - red, \textit{Sadness} - blue, \textit{Neutral} - cyan, \textit{Happiness} - green. We can observe that  \textit{Fine-tuning MP-2 and MP-3} networks can separate \textit{Anger} and \textit{Sadness} classes even though these representations are directly computed from the ASR network without any emotion-specific training. \textit{Fine-tuning RNN} networks benefit from being trained directly for emotion recognition and form visually distinguishable clusters.}
  \label{fig:tsneasrmean}
\end{figure*}

\subsection{Data}

\subsubsection{Emotion data} 

\par The Interactive Emotional Dyadic Motion Capture dataset IEMOCAP  \cite{busso_iemocap:_2008} contains five recorded sessions of conversations between two actors, one from each gender. The total amount of data is 12 hours of audio-visual information from ten speakers annotated with categorical emotion labels (Anger, Happiness, Sadness, Neutral, Surprise, Fear, Frustration and Excited), and dimensional labels (values of the activation and valence from 1 to 5). Similarly as in previous work \cite{huang_attention_2016}, we merged the \textit{Excited} class with \textit{Happiness}. We performed several data filtering steps: we kept samples where at least two annotators agreed on the emotion label, discarded samples where an utterance was annotated with 3 different emotions and used samples annotated with neutral, angry, happy and sad, resulting in 6,416 samples (1,104 of Anger, 2,496 of Happiness, 1,752 of Neutral and 1,064 of Sadness). We use 4 out of 5 sessions for training and the remaining one for validation and testing (as there are two speakers in the session, one was used for validation and the other for testing).

\subsubsection{Speech data}
We concatenated three datasets to train the ASR model: LibriSpeech, TED-LIUM v2, and VoxForge. LibriSpeech \cite{panayotov_librispeech:_2015} contains around 1,000 hours of English-read speech from audiobooks. TED-LIUM v2 \cite{rousseau2014enhancing} is a dataset composed of transcribed TED talks, containing 200 hours of speech and 1495 speakers. VoxForge is an open-source collection of transcribed recordings collected using crowd-sourcing. We downloaded all English recordings\footnote{\url{http://www.repository.voxforge1.org/downloads/SpeechCorpus/Trunk/Audio/Original/48kHz_16bit/}}, which is around 100 hours of speech.
Overall, 384,547 utterances containing ~1,300 hours of speech from more than 3,000 speakers were used to train the ASR model.  We conducted no preprocessing other than the conversion of recordings to WAV format with single channel 32-bit format and a sampling rate of 16,000. Utterances longer than 15 seconds were filtered out due to GPU memory constraints.

\subsection{Extracted features}

The ASR network was trained on power spectrograms with filter banks computed
over windows of 20ms width and 10ms stride. For the progressive network, we used the same features as for the ASR branch, and for the SER-specific branch, we used 13 MFCC coefficients and their deltas extracted over windows of 20ms width and 10ms stride. We have extracted pitch values smoothed with moving average with a window size of 15 using the OpenSMILE toolkit \cite{eyben2013recent}. The reason for the choice of high-level features like MFCC and pitch for the SER-branch was the limited size of emotion annotated data, as learning efficient emotion representation from low-level features like raw waveform or power-spectrograms might be difficult on the dataset of a size of IEMOCAP. In addition, such feature set showed state-of-the-art results \cite{huang_attention_2016}. In both cases, we normalized each feature by subtracting the mean and dividing by standard deviation per utterance.

\begin{figure*}
  
  \centering
  \includegraphics[width=15cm, height=5.5cm]{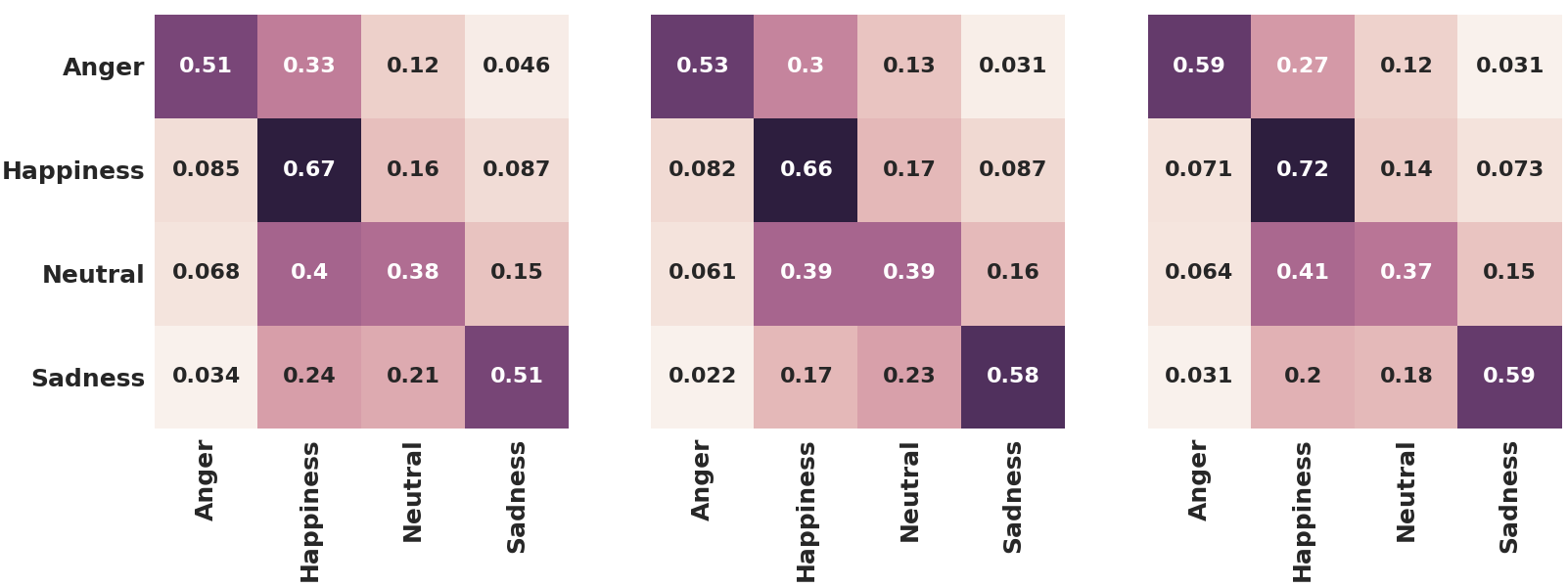}
  \caption{Confusion matrices for our best baseline, \textit{Fine-tuning MP} and \textit{Progressive} models averaged over 10-folds on the IEMOCAP dataset. }
  \label{fig:conf_matrices}
\end{figure*}

\begin{table*}[t]
\centering
\caption{Utterance-level emotion 4-way classification performance (unweighted and weighted accuracy and f1-score). Several variants of fine-tuning and progressive networks are evaluated: using first, second or third ASR layers as input for Fine-tuning MP, Fine-tuning RNN, and progressive networks.}
\bigskip
\begin{tabular}{l*{3}{c}}

\textbf{Model}              & \textbf{U-Acc} & \textbf{W-acc}  & \textbf{F1-score} \\
\hline
Most frequent class & $0.39 \pm 0.04$ & $0.33 \pm 0.2$ &  NaN  \\

Random prediction & $0.25 \pm 0.02$ & $0.25 \pm 0.02$ & $0.24 \pm 0.02$  \\
\hline
\textbf{Baseline 64 units}        &     $0.52 \pm 0.04$  & $0.54 \pm 0.06$ & $ 0.48 \pm 0.05 $  \\
\textbf{Baseline 96  units}        &     \textbf{0.53} $\pm$ 0.01  & \textbf{0.55} $\pm$ 0.03 & \textbf{0.51} $\pm$ 0.02  \\
\textbf{Baseline 128  units}        &     $0.50 \pm 0.04$  & $0.51 \pm 0.06$ & $ 0.47 \pm 0.05 $  \\
\hline
\textbf{Fine-tuning MP-1}        &     $ 0.46 \pm 0.05 $  & $ 0.52 \pm 0.07 $ & $ 0.31 \pm 0.11 $ \\
\textbf{Fine-tuning MP-2}        &     $ 0.54 \pm 0.04 $  & $ 0.54 \pm 0.06 $ & $ 0.52 \pm 0.04 $  \\
\textbf{Fine-tuning MP-3}        &     \textbf{0.55} $\pm$ 0.02  & \textbf{0.56} $\pm$ 0.03 & \textbf{0.53} $\pm$ 0.03  \\
\hline
\textbf{Fine-tuning RNN-1}        &     $0.53 \pm 0.04$  & $0.57 \pm 0.04$ & $0.48 \pm 0.08$  \\
\textbf{Fine-tuning RNN-2}        &     \textbf{0.57}  $\pm$  0.03  & \textbf{0.59} $\pm$  0.05 & \textbf{0.56}  $\pm$  0.03  \\
\textbf{Fine-tuning RNN-3}        &     $0.56  \pm  0.02$  & $ 0.57 \pm  0.04 $ & $0.55  \pm  0.03$  \\
\hline
\textbf{Progressive net-1}        &     $0.56 \pm  0.02$  & $ 0.57 \pm  0.04 $ & $0.55  \pm  0.03$  \\
\textbf{Progressive net-2}        &     \textbf{0.58} $\pm$  0.03 & \textbf{0.61} $\pm$ 0.04  & \textbf{0.57}  $\pm$ 0.03   \\
\textbf{Progressive net-3}        &     $0.57  \pm  0.03$  & $ 0.59 \pm  0.03 $ & $0.56  \pm  0.04$ \\
\end{tabular}

\end{table*}

\subsubsection{Training}
The Adam optimizer \cite{kingma_adam:_2014} was used in all experiments with a learning rate of 0.0001, clipping the norm of the gradient at the level of 15 with a batch size of 64. During the training, we applied learning rate annealing if the results on the validation set did not improve for two epochs and stopped it when the learning rate reaches the value of 1e-6. We applied the SortaGrad algorithm \cite{amodei_deep_2015} during the first epoch by sorting utterances by the duration \cite{hannun_deep_2014}.  We performed data augmentation during the training phase by randomly changing tempo and gain of the utterance within the range of 0.85 to 1.15 and -3 to +6 dB respectively. The model with the lowest cross-entropy loss on the validation set was picked to evaluate the test set performance.

\section{Results}

Table 1 summarizes the results obtained from ASR-SER transfer learning.  We evaluate several baseline models by varying the number of GRU units in a network, and three variants for \textit{Fine-tuning MP-$x$}, \textit{Fine-tuning RNN-$x$} and \textit{Progressive net-$x$} by utilizing representations of layer $x$ of the ASR network. 
We report weighted and unweighted accuracy and f1-score to reflect imbalanced classes. These metrics were averaged over ten runs of a ten-fold leave-one-speaker-out cross-validation to monitor an effect of random initialization of a neural network. Also, our results reveal the difficulty of separating \textit{Anger} and \textit{Happiness} classes, and \textit{Neutral} and \textit{Happiness} (see Figure \ref{fig:conf_matrices}). Our best \textit{Fine-tuning MP-3} model achieved 55\% unweighted and 56\%  weighted accuracy, which significantly outperforms  the baseline (p-value $\leq 0.03$) end-to-end 2-layer GRU neural network similar to \cite{huang_attention_2016} and \cite{ghosh_representation_2016}. The fine-tuning model has around 30 times less trainable parameters (as only the softmax layer is trained) and achieves significantly better performance than the baseline. These results show that putting an additional GRU layer on top of ASR representations affects the performance positively and shows significantly better results than the baseline (p-value $\leq 0.0007$). Results prove our hypothesis that intermediate features extracted by the ASR network contain useful information for emotion classification. 

\par The progressive network consistently outperforms baseline end-to-end models, reaching 58\% unweighted and 61\%  weighted accuracy. In all variants, the addition of the second recurrent layer representations of the ASR's network contributes positively to the performance compared to the baseline. Our results support the hypothesis that the progressive architecture of the network allows to combine the ASR low-level representations with the SER-specific ones and achieve the best accuracy result.
\par
\begin{figure}
  
  \centering
  \includegraphics[width=\linewidth]{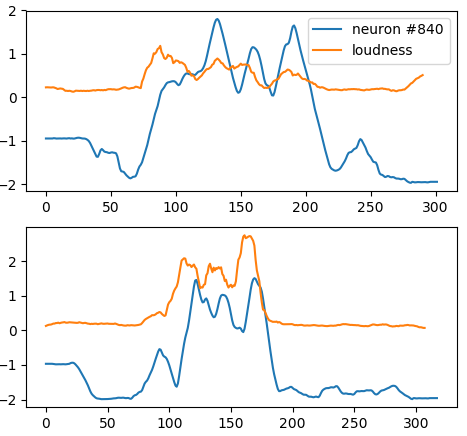}
  \caption{Activations of the neuron number 840 of the second GRU layer of the ASR model (blue) and loudness of two utterances selected randomly from IEMOCAP dataset (orange).}
  \label{fig:loudness_0}
\end{figure}

 In addition to the quantitative results, we tried to analyze the reason of such effectiveness of the ASR representations, by visualizing the representations of the utterances by \textit{Fine-tuning MP-$x$} and  \textit{Fine-tuning RNN-$x$} networks (see Figure \ref{fig:tsneasrmean}). We observe that a prior ASR-trained network can separate \textit{Sadness} and \textit{Anger} samples even by pooling representations of the first ASR layer. On the 3rd layer, \textit{Anger}, \textit{Sadness} and \textit{Happiness} form visually distinguishable  clusters which could explain the surprising effectiveness of \textit{Fine-tuning MP-2/3} models. \textit{Fine-tuning RNN-$x$} networks can separate four classes better due to an additional trained GRU network on top of the ASR representations. Also, we found that activations of some neurons in the ASR network correlate significantly with the well-known prosodic features like loudness. Figure \ref{fig:loudness_0} shows the activation of the neuron number 840 of the second GRU layer of the ASR network and the loudness value of the speech frame for two audio files. We found that, on average, the Pearson's correlation between loudness and activation of the 840th neuron calculated on the IEMOCAP dataset is greater than 0.64 which is an indicator that the ASR network is capable of learning prosodic features which is useful for emotion classification.

\section{Discussion and conclusion}

In this paper, various neural architectures were proposed utilizing speech recognition representations. Fine-tuning provides an ability to use the ASR network for the emotion recognition task quickly. A progressive network allows to combine speech and emotion representations and train them in parallel. Our experimental results confirm that trained speech representations, even though expected to be agnostic to a speaker's emotion, contain useful information for affective state predictions.\par

A possible future research direction would be to investigate the influence of linguistic knowledge on the speech representations and how it affects the system performance. Additionally, the ASR system can be fine-tuned in parallel with the emotion branch by updating the layers of the ASR network. Potentially, this could help the system to adapt better to the particular speakers and their emotion expression style. Furthermore, analyzing linguistic information of the spoken text produced by the ASR network could possibly alleviate the difficulty of separating \textit{Anger} and \textit{Happiness} classes, and \textit{Neutral} and \textit{Happiness}.

\section*{Acknowledgments}
This project has received funding from the European Union's Horizon 2020 research and innovation programme under the Marie Sklodowska-Curie grant agreement No 642667 (SECURE) and partial support from the German Research Foundation DFG under project CML (TRR 169).

\def\url#1{}
\def\href#1{}
\bibliography{Zotero,ijcnlp2017}
\bibliographystyle{ijcnlp2017}

\end{document}